\documentclass{article}

\usepackage{microtype}
\usepackage{graphicx}
\usepackage{subfigure}
\usepackage{booktabs} % for professional tables

\usepackage{hyperref}

\usepackage{url}
\usepackage{relsize}
\usepackage[svgnames]{xcolor}
\hypersetup{colorlinks,urlcolor=Blue}

\usepackage[accepted]{icml2019}

\icmltitlerunning{Representation Learning for Electronic Health Records}

\begin{document}

\twocolumn[
\icmltitle{Representation Learning for Electronic Health Records}




\begin{icmlauthorlist}
\icmlauthor{Wei-Hung Weng}{goo}
\icmlauthor{Peter Szolovits}{goo}
\end{icmlauthorlist}

\icmlaffiliation{goo}{MIT CSAIL, Cambridge, MA, USA}

\icmlcorrespondingauthor{Wei-Hung Weng}{ckbjimmy@mit.edu}

\icmlkeywords{Machine Learning, Clinical, Electronic Health Record, Representation Learning}

\vskip 0.3in
]



\printAffiliationsAndNotice{}  % leave blank if no need to mention equal contribution

\begin{abstract}
Information in electronic health records (EHR), such as clinical narratives, examination reports, lab measurements, demographics, and other patient encounter entries, can be transformed into appropriate data representations that can be used for downstream clinical machine learning tasks using representation learning. 
Learning better representations is critical to improve the performance of downstream tasks.
Due to the advances in machine learning, we now can learn better and meaningful representations from EHR through disentangling the underlying factors inside data and distilling large amounts of information and knowledge from heterogeneous EHR sources. 

In this chapter, we first introduce the background of learning representations and reasons why we need good EHR representations in machine learning for medicine and healthcare in Section 1.
Next, we explain the commonly-used machine learning and evaluation methods for representation learning using a deep learning approach in Section 2. 
Following that, we review recent related studies of learning patient state representation from EHR for clinical machine learning tasks in Section 3.
Finally, in Section 4 we discuss more techniques, studies, and challenges for learning natural language representations when free texts, such as clinical notes, examination reports, or biomedical literature are used.
We also discuss challenges and opportunities in these rapidly growing research fields.
\end{abstract}

\section{Learning Representations for Medicine and Healthcare}
Medicine and healthcare has become one of the key applied machine learning research domains due to increasing adoption of electronic health records (EHR) and the increasing power of computation~\cite{charles2013adoption,topol2019high}. 
Researchers have framed various medical and healthcare-related challenges as machine learning tasks and adopted various algorithms to tackle them with massive amounts of medical data~\cite{topol2019high}. 
Some examples of commonly-seen topics in machine learning for medicine and healthcare research include diagnosis support~\cite{lipton2015learning,choi2016retain,gulshan2016development,esteva2017dermatologist}, outcome and risk prediction~\cite{ghassemi2014unfolding,futoma2015comparison,choi2016doctor,xiao2018readmission}, patient phenotyping~\cite{miotto2016deep,baytas2017patient}, optimal decision making~\cite{raghu2017continuous,weng2017representation,komorowski2018artificial}, and workflow improvement~\cite{horng2017creating,chen2018microscope}. 
Researchers have utilized various types of EHR data in addressing these tasks, such as lab measurements~\cite{pivovarov2015learning}, claims data~\cite{doshi2014comorbidity,pivovarov2015learning,choi2016learning}, clinical narratives~\cite{pivovarov2015learning,weng2017medical}, medical images~\cite{gulshan2016development,esteva2017dermatologist,bejnordi2017deep,liu2017detecting,poplin2018prediction,nagpal2018development}, as well as waveform signals~\cite{lehman2018representation}. Many efforts use multiple such modalities of available data.

For medical and healthcare applications, it is critical to develop robust techniques that can not only yield good performance on given tasks but also provide efficiency, reliability, and explainability~\cite{szolovits1978categorical,szolovits1982artificial}, to improve the likelihood of their practical clinical deployment~\cite{chen2019develop}.
For example, applying an attention mechanism or interpretable models give us better explainability of the model behavior or the prediction~\cite{bahdanau2014neural,ribeiro2016should,lundberg2017unified}.
Designing models with a robust optimization to tolerate adversarial examples provides the model reliability~\cite{madry2017towards}.
Preprocessing data appropriately and making better data representations for algorithms allow us to develop models with better performance and also interpretability. 

A good representation organizes the data in a way that machine learning algorithms can learn models with good performance from them.
It also transforms the data into a form that provides human interpretability given a suitable model design.
For example, the radial domain folding algorithm, an unsupervised multivariate clustering method developed by~\cite{joshi2012prognostic}, abstracts the patient states and summarizes the patient physiology from vitals, labs, and clinical categorical data to a dense but rich representation using domain knowledge.
The resulting model outperforms classical clinical scoring systems on the critical patient mortality prediction task while retaining human understandability of the representation.
A good representation may also be derived from multimodal data sources~\cite{weng2019multimodal}.
~\cite{suresh2017clinical} preprocessed, transformed, and represented the raw data from different modalities (static variables such as demographics, time-varying variables like vital signs and labs, and clinical narrative notes) into a representation for clinical intervention prediction tasks.
They transformed the clinical notes into a low-dimensional vector of topic distributions to preserve the human interpretability of the representation.
Therefore, having appropriate representations is essential for modeling since it provides the fundamental organization of the data in both a machine and human understandable language~\cite{bengio2013representation}. 

\subsection{From Expert-curated to Learning-based Representations}
From domain expert-curated to fully-automated approaches, healthcare information can be represented by different techniques for downstream modeling. 
Traditionally, researchers work heavily with medical experts on data preprocessing and curation to obtain meaningful feature sets (clinical variables) that are derived from domain knowledge. 
The commonly-used clinical scoring and grading models heavily rely on domain experts to identify and curate important clinical features. 
For example, APACHE (Acute Physiology and Chronic Health Evaluation) score~\cite{knaus1985apache} and SOFA (Sequential Organ Failure Assessment) score~\cite{vincent1996sofa} for evaluating the patient severity in the intensive care unit (ICU) setting, CHADS2 score for stroke risk assessment in patients with atrial fibrillation~\cite{gage2001validation}, MELD (Model For End-stage Liver Disease) score for liver transplant~\cite{kamath2001model}, and KDIGO (Kidney Disease Improving Global Outcomes) score~\cite{levey2005definition} for outcome prediction of acute kidney injury, are all scoring models with a small number of clinical predictors that are identified by domain experts or even randomized controlled trials.
The advantage of such expert-intensive feature engineering is interpretability and explainability of developed models. 
However, such an approach is hard to generalize and scale up, and may not be able to capture hidden patterns inside complicated, heterogeneous data. 

Instead of using problem-specific, manually curated predictors, adopting machine learning, specifically representation learning techniques, may help to represent data better for model development, with the potential to discover hidden patterns and new knowledge~\cite{davis1993knowledge,caruana2015intelligible}.
Through learning from data, machine learning techniques allow us to perform feature engineering with less expert effort.
Taking clinical narratives as an example, we can learn the simple but powerful statistics-based lexical features identified by natural language processing (NLP) techniques---bag-of-words or n-gram algorithms---to represent the unstructured free text in a machine understandable form for further model development without annotations from experts~\cite{marafino2014research,weng2017medical}. 
It is also possible to integrate domain knowledge from existing knowledge bases, i.e., ontologies, while learning representations. 
We can utilize the general biomedical knowledge base, Unified Medical Language System (UMLS) Metathesaurus~\cite{bodenreider2004unified}, the Semantic Network~\cite{mccray2001aggregating,mccray2003upper}, Medical Subject Headings (MeSH), or other biomedical knowledge bases to identify and transform the unstructured medical information into meaningful and machine-comprehensible representations.
Clinical NLP systems such as Apache clinical Text Analysis and Knowledge Extraction System (cTAKES)~\cite{savova2010mayo}, MetaMap~\cite{aronson2001effective}, Clinical Language Annotation, Modeling and Processing Toolkit (CLAMP)~\cite{soysal2017clamp}, and the Clinical Named Entity Recognition system (CliNER)~\cite{boag2018cliner}, help annotate and extract clinically meaningful concepts from unstructured clinical texts and link them to standard terminologies, such as UMLS concept unique identifiers (CUIs), to obtain interpretable representations in a unified language.
For example,~\cite{weng2017medical} applied both simple lexical features and identified CUIs with semantic filtering to unstructured clinical free texts and obtained clinically meaningful representations for a downstream classification task. 
They demonstrated that the representation considering both language (lexical features) and domain knowledge (CUIs) outperformed other methods on the classification task, with clinical interpretability.
The standardization of representation using ontology also provides a great opportunity to harmonize unstructured data in multiple datasets, which is essential for developing models across datasets~\cite{gong2017predicting}.

Learning representations directly from raw data without extracting and mapping concepts to existing knowledge bases is an alternative to further reduce human involvement, which may increase the machine's potential for exploring and identifying hidden patterns inside the data. 
~\cite{fonarow2005risk} utilized the classification and regression trees algorithm (CART) to automatically create a series of clinically meaningful rules purely from data, and applied the rules to the mortality risk stratification problem for patients with acute decompensated heart failure. 
We can also learn high-level abstract representations by disentangling the underlying factors and distilling large amounts of information in heterogeneous clinical data through advanced learning algorithms.
Such abstract representations usually provide generalizable power for different machine learning scenarios such as semi-supervised learning, multitask learning~\cite{weng2019multimodal}, transfer learning, and domain adaptation, which is useful for common medical and healthcare-related tasks where data are scarce or inaccessible for an intended application but similar data, say from other institutions, are more easily available. 
~\cite{ghassemi2014unfolding} applied the topic modeling algorithm, latent Dirichlet allocation (LDA)~\cite{blei2003latent}, to learn the latent representations of clinical progress notes and predict mortality in the critical care setting. 
For time-series data, the hyper-parameters used for non-parametric multitask Gaussian processes can be the latent features~\cite{ghassemi2015multivariate}, and switching-state autoregressive models can also model the underlying state representations~\cite{ghassemi2017predicting}. 

Among various machine learning algorithms, deep learning---a family of neural network-based algorithms advocated by connectionism~\cite{rumelhart1986parallel}---is the one that can learn more abstract representations through multiple non-linear transformations within a highly modularized framework~\cite{bengio2013representation,lecun2015deep}.
Learning representations using deep learning has achieved numerous successes in several domains with different data modalities, such as natural language in free text~\cite{mikolov2013distributed,mikolov2013efficient,le2014distributed,bojanowski2016enriching,peters2018deep,devlin2018bert}, audio and speech processing~\cite{chung2016audio,chung2018speech2vec}, and computer vision~\cite{krizhevsky2012imagenet}. 
Researchers in the medicine and healthcare domain are also making great efforts to approach problems using a deep learning approach~\cite{gulshan2016development,yala2017using,chung2017learning,raghu2017continuous,choi2016doctor}.

\section{Basics of Deep Learning-based Representation Learning}
Representation learning is one of the successful and exciting fields in recent machine learning research. 
The hallmark of contemporary machine learning has been to transform discrete problems and representations into continuous ones, where models are typically differentiable and therefore continuous optimization techniques---rather than combinatorial discrete methods---can be applied.

The goal of representation learning is to encode and represent (embed) raw input information into small, dense and distributed embedding vectors (embeddings) in a continuous vector space, where similar inputs can be mapped to nearby points, i.e. embedded close to each other.
The inputs can be either dense or sparse, e.g., image pixels, audio segments, time points in time-series or irregularly occurring events, numbers, words, context, or clinical concepts, depending on the task, and the embeddings are usually computed via optimizing the parameters of neural network models given machine learning tasks, such as classification, regression, sequence prediction, next word prediction, and the corresponding objective functions.
Such learned embeddings may capture semantic, linguistic, temporal or spatial relations between the inputs~\cite{mikolov2013distributed,mikolov2013efficient,krizhevsky2012imagenet}, computed from statistical properties of the relations among the data.
These representations can be used not only directly for similar information retrieval, such as a search for similar clinical cases, but also as inputs for various downstream machine learning tasks, such as clinical prediction and classification, due to their generalizability~\cite{chung2017learning,weng2017representation}. 
Compared to traditional discrete encoding techniques such as one-hot encoding and the bag-of-words model, the distributed representation learning approach better handles the issues of the curse of dimensionality, matrix sparsity, and feature dependency in the discrete encoding approach since the features can be embedded in a low-dimensional space.

\subsection{Mechanisms}
Due to advances in neural network architecture design, we now can embed the raw input into a vector space using different network architectures based on characteristics of given tasks. 
For example, we use convolutional neural network (CNN)-based models, such as AlexNet, Inception, or ResNet, for computer vision problems to preserve spatial information~\cite{lecun1998gradient}, recurrent neural network (RNN)-based models---conventional RNN, RNN with long short-term memory (LSTM) or gated recurrent units (GRU)~\cite{hochreiter1997long,cho2014properties}, and attention-based models like Transformer for free texts and time-series data to keep more sequential properties~\cite{vaswani2017attention,devlin2018bert}.
We may also adopt graph-based models for data with underlying network structure~\cite{kipf2016semi}, or just simply use multilayer perceptrons (MLP) if the features are independent and identically distributed. 

The function of neural networks can further be augmented by adding various techniques for different purposes, such as sequence-to-sequence (seq2seq) and encoder-decoder frameworks for learning structured information~\cite{sutskever2014sequence,cho2014properties,cho2014learning}, attention mechanisms for model interpretability~\cite{bahdanau2014neural,vaswani2017attention}, or generative adversarial network (GAN) for data synthesis~\cite{goodfellow2014generative}. 

\paragraph{Encoder-Decoder Architecture}
There are some fundamental neural network designs for representation learning with the deep learning approach.
The most intuitive and general architecture is the encoder-decoder framework design, such as a sequence-to-sequence (seq2seq) or autoencoder model~\cite{hinton2006reducing,sutskever2014sequence}. 
The encoder is simply a function that maps an input space to a latent space, and the decoder is another function that maps the latent space to a target space.
One can design an encoder-decoder system using any neural network components, such as CNN or RNN, to encode the complex input into a compressed latent space representation, and decode the representation to a target output.
The latent representations, i.e., embeddings, in the network layers between the encoder and decoder can therefore become the representation of the given input. 

For example, seq2seq is a framework that takes a sequence of input data and transforms it into a latent representation, and decodes the representation to another sequence, such as in the task of English to French translation. 
Autoencoders form a family of neural network-based models---a special case of an encoder-decoder framework---to learn to transform a complex input into a compressed representation and then to translate that into a reconstruction of the input, minimizing reconstruction loss, i.e., the difference between the model input and output~\cite{rumelhart1986learning,hinton2006reducing}. 
Therefore, the decoder in an autoencoder framework is a structure that reverses the functionality of the encoder. 
For example, doing convolution while encoding and deconvolution while decoding using CNN as the components.
Since the output and input for an autoencoder framework are the same, the method therefore become a common approach in unsupervised learning scenarios. 
The intuition behind this form of representation learning is that the compressed representation best captures the essence of the data while ignoring happenstantial variation.

Researchers also apply sparse autoencoders (SAE) and denoising autoencoders (DAE) to learn sparse representations and learn robust representations through noise injection~\cite{vincent2008extracting}. 
Such variants of autoencoder are helpful when data are noisy or has significant missingness---for example, vital signs and lab measurements in the EHR.

To obtain task-specific representations for predictive models, the decoder can be replaced by other network components, such as fully connected layers, with an appropriately defined loss function (objective function), whose aim is to optimize the model via specific downstream auxiliary tasks.
In this case, the learned representation will be biased to the given task. 
Using downstream auxiliary tasks is a common approach to learn patient state representations from EHR.
We will discuss this more in Section 4.

\paragraph{Learning Representations from Sequences}
Another mechanism of representation learning is to learn co-occurrence information from sparse inputs or sequences, such as sentences, paragraphs, documents, or time-series signal sequences. 
For example, two popular models in NLP belong to this category---word2vec and GloVe (Global Vectors)~\cite{mikolov2013distributed,mikolov2013efficient,pennington2014glove}.
The word2vec model uses unsupervised skip-gram and continuous bag-of-word (CBOW) algorithms to obtain the embedding in a vector space of the tokens in an online fashion. 
Skip-gram' objective is, for each word $w(n)$, to minimize the difference of predicted and actual probabilities of tokens $\{w_{n-k}, ..., w_{n-1}, w_{n+1}, ..., w_{n+k}\}$ within a window of size $k$ of $w(n)$. 
The objective of CBOW, on the other hand, aims to infer the current token $w(n)$ from its nearby tokens $\{w_{n-k}, ..., w_{n-1}, w_{n+1}, ..., w_{n+k}\}$~\cite{mikolov2013distributed}.
In either case, a single-layer neural network is trained to optimize these predictions, and its weight vector is taken to be the embedding of $w(n)$.
The GloVe model instead aims to learn the embedding space via precomputing the co-occurrence matrix of the whole corpus and factorizing it using the asynchronous stochastic gradient descent (SGD) algorithm~\cite{pennington2014glove}. 
These methods can also be generalized to any sequential data modalities.

Word-level representations can also be learned at the subword level, i.e., using character information~\cite{bojanowski2016enriching}. 
The word representation with subword information may overcome the issue of out-of-vocabulary or misspelled words and may better capture the semantics in word morphology~\cite{bojanowski2016enriching,weng2019unsupervised}.
Many of medicine and healthcare NLP studies use either word2vec or GloVe to train the word or concept-level representations or adopt pre-trained word2vec or GloVe embeddings for their downstream tasks~\cite{wang2018comparison}.
However, the limitation of such approaches is that they determine a single embedding vector for all occurrences of a token and do not consider the context around the token, which may lead to the issue of word sense ambiguity.

\paragraph{Model Pre-training and Transfer Learning}
Learning representations from pre-trained models trained on large datasets, such as ImageNet, and using the features extracted from such models to modify the weights of a target model (fine-tuning), often called transfer learning, has been a relatively standard approach for computer vision tasks~\cite{krizhevsky2012imagenet}.

Recently, model pre-training has also dramatically reshaped the NLP community due to its ability to capture better semantics~\cite{howard2018universal} and allow transfer learning from large general domain language corpora to smaller domain-specific NLP tasks~\cite{devlin2018bert,howard2018universal,yang2019xlnet}.
With large corpora, context-aware models such as ELMo (Embeddings from Language Model)~\cite{peters2018deep}, ULMFiT (Universal Language Model Fine-tuning)~\cite{howard2018universal}, GPT (Generative Pre-training Transformer)~\cite{radford2018improving}, BERT (Bidirectional Encoder Representations from Transformers)~\cite{devlin2018bert}, and XLNet~\cite{yang2019xlnet} can pre-train the language model in an unsupervised way and fine-tune the model using task-specific supervised learning with auxiliary tasks.
% kiros2015skip,logeswaran2018efficient,conneau2017supervised

ELMo concatenates independently trained multi-layer LSTMs in two directions to learn a contextualized representation without supervision~\cite{peters2018deep}, ULMFiT first incorporates the ideas of pre-trained language model and fine-tuning for transfer learning in NLP~\cite{howard2018universal}, GPT adopts the multi-layer transformer decoder as language model~\cite{radford2018improving}, BERT also takes advantages of a multi-layer transformer but uses it as a bi-directional encoder to acquire the natural language representation via two general auxiliary language tasks---masked language model (MLM) and next sentence prediction~\cite{devlin2018bert}. 
Differently from BERT, which trains the model in a denoising autoencoder fashion with MLM, XLNet uses an autoregressive pre-training method to obtain representations that yield even better performance on multiple natural language benchmark tests~\cite{yang2019xlnet}.
ULMFiT, GPT, BERT and XLNet are agnostic to downstream tasks and therefore very flexible for transfer learning in natural language problems.

Such a pre-trained and fine-tuning framework and a contextualized learning scheme mitigates the issue of word sense ambiguity by considering the surrounding context in general language models, which is critical in medical and healthcare domains, and has become the main approach for learning the natural language representation for all kinds of NLP tasks.

\subsection{Evaluation}
The learned representations, or embeddings, can be evaluated in both quantitative and qualitative ways. 
For quantitative evaluation, downstream tasks are usually required.
For example, we can conduct the information retrieval task to identify the most similar cases or tokens given learned representations and queries~\cite{wang2018comparison,weng2018mapping,hsu2018unsupervised}. 
Such tasks can be evaluated by accuracy, precision at $K$, or other metrics for information retrieval---mean reciprocal rank (MRR), mean average precision (MAP), and normalized discounted cumulative gain (nDCG)~\cite{jarvelin2002cumulated}. 
The quality of learned representations can also be evaluated by specific auxiliary measures based on the performance of the task.
For instance, studies use accuracy, the area under the ROC curve (AUROC, or simply AUC), precision, recall, and F1-score for prediction tasks~\cite{miotto2016deep,weng2017medical}, area under the Precision-Recall Curve (PR-AUC) for prediction tasks with imbalanced data~\cite{choi2018mime}, or other task-specific metrics such as BLEU score for NLP tasks.

The qualitative evaluation of the learned representations can be done by retrieval of similar cases~\cite{weng2018mapping,hsu2018unsupervised,weng2019unsupervised} or visualization~\cite{choi2016multi,chung2017learning,xiao2018readmission}. 
Similar case retrieval is an interpretable, case-based reasoning approach to evaluate the quality of representations. 
For example,~\cite{weng2018mapping,weng2019unsupervised} listed a few queries with their closest neighbors in the embedding vector space to demonstrate that the learned representations do capture the semantics of the corpora. 

Visualization instead requires dimensionality reduction algorithms to get the dimensions to be two or three, which is visualizable for human interpretation.
Principal component analysis (PCA) is a linear method for visualization that finds the principal components of the data by transforming data points into a new coordinate system. 
The non-linear algorithm, t-Distributed Stochastic Neighbor Embedding (t-SNE), is an alternative when we want to explore or visualize the data with higher dimension~\cite{maaten2008visualizing}. 
t-SNE is able to map high-dimensional data into a low-dimensional manifold by creating an embedding that attempts to maintain local structure within the data. 
However, t-SNE itself cannot be a method to learn representations since the model doesn't retain distances but estimates probabilities. 
% t-SNE assumes and maps a high-dimensional space with Gaussian distribution to a low-dimension sapce with t-distribution, and uses Kullback-Liebler (KL) divergence to evaluate the loss between between two distributions.
% Thus, errors between the Euclidean distances become useless.
Instead, Uniform Manifold Approximation and Projection (UMAP) is another manifold learning model that does support transforming new data into the embedding vector space, which allows UMAP to perform representation learning~\cite{mcinnes2018umap}. 
Google researchers provide Embedding Projector for visualizing the learned representation using different models.\footnote{\url{https://projector.tensorflow.org/}.}

\section{Learning Patient State Representations}
Learning good patient state representations is a critical step in clinical machine learning before conducting downstream tasks since the raw clinical data are usually unstructured and sparse~\cite{hripcsak2012next}. 
It is helpful to discover cohort or disease phenotypes and to predict the outcome of interest through the process of pattern mining of given heterogeneous medical and healthcare data. 
To learn patient state representations, researchers usually leverage supervised downstream tasks for optimization, such as classification or regression, although unsupervised settings may provide less performant but more generalizable representations.
Researchers can take advantage of modularized neural network architectures to develop such end-to-end learning scenarios. 
For example, in the Deep Patient model,~\cite{miotto2016deep} used a three-layer stacked DAE with sigmoid activation functions to encode patient representations using diagnoses, medications, procedures, lab test codes, and 300 LDA-transformed clinical note topics, from the longitudinal EHR of 704,857 patients. 
They evaluated the model through disease classification and patient disease tagging tasks using random forest classifiers. 
The results show that the deep patient representations yielded better predictions than raw EHR features and the representations learned from PCA and $k$-means algorithms.

In this section, we discuss several issues and challenges while learning effective patient state representations, such as modeling EHR temporality and time irregularity found in patient histories and hospital visits, modeling the EHR hierarchy, domain knowledge injection, and model interpretability. We also review recent related studies about patient state representation learning.

\subsection{Temporality and Irregularity}
The temporal information in EHRs is necessary for learning better patient state representations since health status and disease both progress over time. 
Learning time-aware representations is critical to improving the performance of clinical decision support task.
~\cite{che2015deep} utilized a stacked autoencoder to discover patient phenotypes and learn patient-level representations from clinical time-series data. 
Additionally, they integrated the tree-based ICD ontology as prior knowledge to regularize parameters in the top layer of the neural network. 
Such prior-based regularization biases the model toward prior domain knowledge and benefits performance.
~\cite{lasko2013computational} also used a stacked autoencoder on time-series uric acid measurements to classify gout and acute leukemia. 
Researchers used CNN- and RNN-based neural networks to tackle the complex temporality of medical events in EHR as well.
~\cite{cheng2016risk} adopted the CNN-based model for the issue of temporality. 
They used a four-layer 1D-CNN with observation and prediction windows to learn the patient representation and predict congestive heart failure and chronic obstructive pulmonary disease.
~\cite{nguyen2017mathtt} proposed the \texttt{Deepr} model, which is also a CNN-based network, to identify predictive clinical motifs in the longitudinal EHR to perform risk prediction and detect interpretable clinical patterns.

Using RNN-based models,~\cite{lipton2015learning} learned the patient representation via an LSTM on sequential lab measurements. 
Their model classified 128 diagnoses with 13 frequently but irregularly sampled measurements from patients in a pediatric ICU and outperformed other strong baseline models such as logistic regression and MLP with expert-curated features. 
In the Doctor AI model,~\cite{choi2016doctor} developed a GRU model to encode the sequential patient history into a patient-visit representation to make the differential diagnosis (multilabel prediction) for the new visit. 
The Doctor AI disease progression model outperformed logistic regression and MLP on the multilabel prediction reported by recall@30. 
They also demonstrated that domain adaptation is possible using the proposed technique. 
The learned coefficients can be transferred as the initialization of a new task on a different dataset.
~\cite{choi2016using} also used the GRU model that encodes the patient visit-representation by aggregating the learned code-level representations in the longitudinal EHR and used the representation to predict heart failure. 
They designed different time intervals of the observation window and the prediction window as the training and testing data, respectively, for outcome prediction. 
Both GRU and time interval window design help the model outperform other methods such as logistic regression, SVM, MLP and $k$-nearest neighbor (KNN).
~\cite{suresh2017clinical} learned the patient representation using either CNN or LSTM with combined features, including static information, time-series lab and vital data, and LDA-transformed unstructured clinical notes.
They compared the performance on five ICU intervention prediction tasks among different model architectures, and showed that given the learned representations, CNN and LSTM network architectures are similarly effective.
The clinical interpretability can also be evaluated by feature-level occlusion in LSTM, or by convolutional filters in CNN.

The issue of time irregularity may also be approached by tweaking the components of neural network architectures. 
The \texttt{DeepCare} model adopted the LSTM, pooling and word embedding to encode patient history, infer current illness and predict outcome~\cite{pham2016deepcare}. 
They used time decay and time parameterization on the forget gate in the LSTM unit to handle time irregularity. 
For each hospital admission, a single vector representation was learned. 
They integrated the intervention information to augment prediction power for disease progression modeling, intervention recommendation and future risk prediction. 
Instead,~\cite{baytas2017patient} proposed a time-aware LSTM (T-LSTM) autoencoder to tackle irregularity of time and learn a patient-level representation given sequential records of a patient. 
The LSTM cell memory learns time decay to discount the memory content according to the elapsed time, which is in an unsupervised learning setting. 
This patient subtyping model does capture the underlying structures in the sequences with time irregularities.
~\cite{che2017rnn} developed a GRU-based 2D-RNN model that applies the concept of dynamic time warping to measure similarity between two temporal sequences to model the gate parameters in the GRU.
These approaches are all useful to tackle the issue of time irregularity.

\subsection{Hierarchy}
Instead of flattening different sets of information in the EHR by simply concatenating all vector representations, utilizing the inherent multilevel structure, such as the diagnosis-treatment relationship, of EHR can be useful to learn the patient state representations.
In the Multilevel Medical Embedding (\texttt{MiME}) model, authors considered the multilevel structure of EHR and the complex interaction between diagnosis and treatment codes to encode the patient representations~\cite{choi2018mime}. 
They used the same auxiliary prediction task as~\cite{choi2016using} and conducted experiments in different settings of data size and visit complexity (disease-treatment interactions). 
The patient-level \texttt{MiME} representations outperformed standard word-level representations~\cite{choi2018mime}, \texttt{med2vec}~\cite{choi2016multi}, and \texttt{GRAM} concept-level representations~\cite{choi2017gram}.
~\cite{choi2019graph} further developed a graph convolutional transformer (GCT) to capture the underlying EHR structure when the underlying explicit structure information is missing.
For example, the information in claims data is usually flat and we have no clues which treatment is related to certain lab data.
GCT can discover implicit underlying structure with a Transformer-based model design.
~\cite{xiao2018readmission} proposed the \texttt{CONTENT} model that hybridizes both the deep neural network and the probabilistic generative model to preserve the long-term (global context) and short-term information (local context), respectively, in the EHR. 
The authors used RNN to capture short-term local context, and topic modeling to learn the long-term global context of a patient's medical history. 
Such time-aware patient representations yield better performance while predicting hospital readmissions. 

\subsection{Domain Knowledge}
As mentioned in the beginning of the chapter, we can leverage the machine learning methods to learn a good representation using domain knowledge~\cite{joshi2012prognostic}.
Unlike many other domains, the clinical and biomedical world has many invaluable biomedical knowledge bases, expert-curated ontologies and a metathesaurus---UMLS---containing SNOMED-CT, ICD, CPT, LOINC, and NDC terminologies to help us while conducting clinical medicine and healthcare machine learning tasks. 
Injecting the prior knowledge leverages the interpretability and robustness of the learned representations. 
One can regard such a process as adding model regularization or injecting a bias toward the experts' ``thoughts'' and human judgment.

\cite{che2015deep} implemented a prior-based graph Laplacian regularization that integrates the relational information in the ICD-9 ontology represented as a weighted graph. 
Yet the graph Laplacian approach requires the appropriate definition of distance on graphs, which is usually not available. 
Instead, in the \texttt{GRAM} model the authors used a knowledge directed acyclic graph (DAG) to encode the ICD-9 ontology into a vector representation that considers its hierarchical relationship~\cite{choi2017gram}. 

\subsection{Interpretability}
Neural network models are notoriously known as black-box methods. 
However, there are actually some approaches that can uncover the black-box and obtain some interpretability of the neural network-based models.
For clinical medicine and the healthcare domain, it is critical to have not only quantitative evaluation but also qualitative, interpretable evaluation to convince medical professionals to adopt the technique. 
Otherwise, machine learning solutions won't be able to turn into deployable and actionable clinical decision support systems even with superior performance.

Interpretability can be considered at three stages---before, during and after model development.
We can use exploratory data analysis (EDA) and data visualization to provide data interpretability before modeling.
During model development, we may also obtain insights from ruled-based models, example-based case-based reasoning, or mimicking the model through knowledge distillation.
Last but not least, we can explain model behavior or predicted results by giving interpretations after model development.
For example, using an attention mechanism or encoder-generator framework to highlight where is the most consequential input for a model's prediction~\cite{bahdanau2014neural,lei2016rationalizing}. 

The attention mechanism has been used in learning patient state representations~\cite{choi2017gram,girkar2018predicting}.
~\cite{choi2016retain} further developed the \texttt{RETAIN} model using a reverse time attention mechanism to mimic clinician behavior by time-reversing the EHR events. 
Such a design means that recent hospital visits are likely to receive higher attention, which may yield clinically actionable results.
Local Interpretable Model-agnostic Explanations (LIME)~\cite{ribeiro2016should}, as well as SHapley Additive exPlanations (SHAP)~\cite{lundberg2017unified}, can both provide the unified framework and visualizable explanations to interpret the model outputs.

\section{Learning Clinical Language Representations}
Clinical narratives such as clinical notes, examination reports, and biomedical literature can also become materials for learning patient state representations.
However, additional techniques and modeling strategies are required due to the unstructured and discrete, sparse nature of natural language.
When building machine learning models based on text in EHRs, one may adapt general purpose NLP approaches or one can focus on using medical terminologies and knowledge bases.

Representations of natural language tokens, such as word, sentence, document, or concept vector representations (embeddings), can be learned from the free texts in EHR by various models for sequences and transfer learning mentioned in Section 2.1, such as word2vec, GloVE, ELMo, or BERT.
These models can embed the discrete language tokens into a continuous vector space as distributed, dense embeddings depending on the distributional hypothesis that argues the words that occur in the same contexts tend to have similar semantics~\cite{harris1954distributional}.
The advantage of learning natural language token embeddings is to obtain generalizable representations for later use.

In the general domain, researchers pre-trained word2vec\footnote{\url{https://code.google.com/archive/p/word2vec/}}, GloVe\footnote{\url{https://nlp.stanford.edu/projects/glove/}}, ELMo\footnote{\url{https://allennlp.org/elmo}} and BERT\footnote{\url{https://github.com/google-research/bert}} embeddings on Wikipedia and Google News corpora.
Such embeddings are widely used in different general and specific natural language tasks as the initial model, whose parameters are then incrementally tuned using smaller but task-specific datasets available from specific domains.

For learning clinical language representations, the type of natural language token (e.g., words, sentences, documents, or medical concept tokens, terms, or phrases) and the source of learning data play important roles in yielding a better quality of representation.
In this section, we discuss the representation learning of natural language tokens and clinical concepts, and the challenges and opportunities of learning language representations such as data insufficiency, domain knowledge injection, and cross-domain/modal resource utilization.

\subsection{Natural Language Tokens Representations}
With the standard representation learning techniques, word-level language representations (word embeddings) have been widely used in various clinical NLP applications such as named entity recognition (NER)~\cite{dernoncourt2017neuroner}, medical synonym extraction, relation extraction (RE), biomedical information retrieval, and abbreviation disambiguation~\cite{wang2018comparison}. 
To learn word-level representations in the medical domain,~\cite{pyysalo2013distributional} used skip-gram to train on PubMed, PMC texts and the Wikipedia corpus to obtain the word embeddings from both biomedical and general-domain large corpora\footnote{\url{http://bio.nlplab.org/}}.
~\cite{minarro2014exploring} also applied skip-gram to multiple biomedical-related corpora, such as PubMed, Merck Manuals~\cite{beers1999merck}, Medscape, Wikipedia, and combined corpora, to learn the word embeddings. 

Sentence, paragraph and document-level representations can also be adopted for developing clinical machine learning models. 
To build higher-level representations, we may use the bag-of-words representation that simply summarizes or averages the word vectors in the sentence, paragraph, or document~\cite{weng2017medical}.
However, such an approach discards the information between tokens in the lower language hierarchy, e.g., the relationship between words.
Instead, we can use contextualized representation learning methods to develop context-aware langauge representations.
For example,~\cite{hsu2018unsupervised} learned the sentence and paragraph representations from the pre-trained embeddings of a Universal Sentence Encoder that further fine-tuned on their own radiology corpus for the radiology report retrieval task.

\paragraph{Data Insufficiency and Transfer Learning}
For learning higher-level language token representations, the techniques based on deep learning and neural networks are data hungry. 
Insufficient data is usually an obstacle while learning representations directly from EHR or conducting any clinical machine learning projects. 
To handle the issue, we may leverage existing knowledge bases, the complex hierarchical structure of given data, and we can take advantage of the transfer learning framework. 
Researchers have used the concept of transfer learning to tackle the problems of medical database encoding change~\cite{gong2017predicting} and hospital-specific prediction tasks~\cite{wiens2014study,weng2017medical}.
~\cite{dubois2017learning} transferred the learned representation from the source task of drug code prediction to the target task of patient phenotype prediction, which provides more potential clinical impact.

Well-learned latent representations of larger corpora can serve as general pre-trained language models for transfer learning across different machine learning tasks. 
We can fine-tune the model by starting with pre-trained language models trained on vast general purpose corpora, and then incrementally fine-tuning these models using the typically smaller data sets available from medical corpora.
~\cite{khin2018deep,zhu2018clinical}\footnote{\url{https://github.com/noc-lab/clinical\_concept\_extraction}} applied pre-trained ELMo to medical texts for de-identification and other clinical NLP benchmark tasks~\cite{uzuner20112010}.
~\cite{lee2019biobert} released the BioBERT model\footnote{\url{https://github.com/dmis-lab/biobert}}, which is trained on a general domain corpus and fine-tuned on biomedical text such as PubMed.
~\cite{alsentzer2019publicly} took one more step toward EHR by pre-training clinically oriented BERT models with clinical notes in the MIMIC-III database~\cite{johnson2016mimic}, either all notes or focusing on discharge summaries, on top of BERT and BioBERT models, and demonstrates that the specialized clinical BERT models outperformed others in the clinical NLP tasks\footnote{\url{https://github.com/EmilyAlsentzer/clinicalBERT}}.
~\cite{huang2019clinicalbert} also developed a clinical BERT model\footnote{\url{http://bit.ly/clinicalbert\_weights}} by fine-tuning the BERT model on EHR for the hospital readmission task.
Such improvement on clinical specific tasks may result from the difference in linguistic features between general, biomedical and clinical narratives.
~\cite{si2019enhancing} investigated the capability of a traditional word- or subword-level approach, e.g., word2vec, GloVe, fastText, and the contextualized methods like ELMo and BERT on a clinical concept extraction task and demonstrated that the contextualized methods achieve better performance on various benchmark tests in the i2b2 and SemEval datasets.

\subsection{Concept Representations}
Biomedical concept representations contain abundant hidden relationships between each of them that cannot be simply represented by one-hot coding or natural language token representations~\cite{choi2016multi}.
For example, when we want to have the concept ``congestive heart failure'' in the embedding space, it can not be represented well simply using the word-level representation, which may represent the phrase using the average of three separate word vector representations for ``congestive'', ``heart'', and ``failure''~\cite{weng2017medical}. 
Instead, learning concept-level representations is an approach to tackle the issue. 
The advantage of concept-level representations is to alleviate the errors that result from word-level representations. 
The key to learning concept-level representations is the process of clinical concept identification and extraction from the clinical narratives.
Next, concept standardization using ontology mapping is needed, which yields a complex training process and the necessity for ontology systems. 
Researchers may need to rely on a clinical concept extraction pipeline, such as cTAKES and MetaMap~\cite{savova2010mayo,aronson2001effective}, for ontology mapping to identify and extract UMLS CUIs from free text.
However, using such concept extraction tools is likely to inject errors into the learning process, so further concept filtering and curation are inevitably needed~\cite{wiegreffe2019clinical}.
%we also need to consider that using the concepts identified by the clinical concept extraction pipeline like cTAKES may not be helpful for all cases, either in data augmentation or multitask perspective, due to the nature of noisiness of cTAKES output~\cite{wiegreffe2019clinical}.
%Further concept filtering and curation is inevitable while using such general pipeline for better model development.

With standard machine learning methods after concept extraction from the free texts,~\cite{de2014medical} derived the embeddings of UMLS CUIs from 348,566 medical journal abstracts using a skip-gram model\footnote{\url{https://github.com/clinicalml/embeddings/raw/master/DeVine\_etal\_200.txt.gz}}.
~\cite{finlayson2014building} learned the UMLS CUIs embedding using 20 million clinical notes spanning 19 years of data from Stanford Hospital and Clinics using co-occurrence based analyses\footnote{\url{https://github.com/clinicalml/embeddings/raw/master/stanford\_cuis\_svd\_300.txt.gz}}.
~\cite{beam2018clinical} also learned the UMLS CUIs embeddings, \texttt{cui2vec}, from medical billing codes, biomedical journal texts, and the clinical concept co-occurence matrix used in~\cite{finlayson2014building}\footnote{\url{http://cui2vec.dbmi.hms.harvard.edu/}}.
~\cite{choi2016learning} learned three dense, low-dimensional embedding spaces of UMLS CUIs and billing codes from UMLS-processed journal abstracts, UMLS-processed clinical notes and claims data using the word2vec skip-gram framework\footnote{\url{https://github.com/clinicalml/embeddings/}}.
~\cite{tran2015learning} used a restricted Boltzmann machine (RBM) to learn abstractions of ICD-10 codes on mental health patients to predict risk of suicide. 

Considering the temporal and hierarchical properties of EHR information, researchers have developed advanced models that consider the more complicated nature of EHR structure to learn concept-level representations efficiently. 
The \texttt{Med2Vec} algorithm is a word2vec variant that learns representations of both code concepts and patient visits from EHR~\cite{choi2016multi}. 
It applies the skip-gram algorithm at a patient visit level, binarizes the diagnosis codes (ICD-9) that are grouped by the Clinical Classifications Software (CCS) grouper, medication codes (National Drug Codes (NDC)) and procedure codes (Category I of Current Procedural Terminology (CPT)) of each visit to learn both code-level and patient visit-level representations.
~\cite{choi2017gram} further proposed a graph-based model with attention mechanism (\texttt{GRAM}) to learn the clinically interpretable hierarchical ICD concept representation that yields better performance on the downstream tasks. 
~\cite{luo2015subgraph} developed a novel framework, subgraph augmented non-negative tensor factorization (SANTF), that converts concepts inside clinical sentences into a graph representation with subgraphs that are clinically interpretable. 
They applied the SANTF model to lymphoma pathology reports and identified the graph and subgraph structure between concepts.
~\cite{mullenbach2018explainable} learned medical code representations by an attentional convolutional network that can predict medical codes from clinical notes, which also provides the explainability of the predicted model output.
~\cite{beaulieu2018learning} adopted the hyperbolic space embedding within the Poincar{\'e} ball~\cite{nickel2017poincare}, to encode the hierarchical property of the ICD ontology. 
The results demonstrated that the hyperbolic space embedding preserves the tree structure of the ICD ontology. 
Yet further investigation is required to see whether the performance and quality of hyperbolic space embeddings are better than Euclidean space embeddings.

\subsection{Sources of Learning Data}
With varying levels of generalizability, representations can be learned from (1) a specific clinical corpus such as clinical notes or articles (e.g., MIMIC-III, i2b2 corpus, Merck Manual), (2) a general biomedical corpus (e.g., biomedical publications in PubMed), or (3) a general natural language corpus such as Wikipedia or Google news. 
The advantage of using a more specific corpus for training is that the learned representations can be highly optimized for the specific domain and purpose. 
For example, using the representations learned from clinical notes yields better performance on a clinical cross-domain translation task than those learned from the PubMed corpus and Wikipedia corpus~\cite{weng2018mapping,weng2019unsupervised}.
~\cite{alsentzer2019publicly,si2019enhancing} also demonstrated that fine-tuning the pre-trained BERT model on an EHR corpus outperformed the model trained on a general corpus and/or biomedical corpus on various clinical machine learning tasks.

~\cite{wang2018comparison} evaluated the performance of using different data sources for different clinical machine learning tasks. 
They concluded that the representations trained on an EHR corpus capture the semantics of clinical terms better and better aligned with experts' judgments than the representations learned from the general natural language corpus, such as the Wikipedia corpus, in a qualitative evaluation. 
However, there is no consensus about which approach is better according to quantitative evaluations. 
Most studies without specific downstream purposes choose to merge various data sources, such as PubMed journal abstracts, clinical notes, claims data, for concept-level representation learning~\cite{de2014medical,finlayson2014building,choi2016learning,beam2018clinical}.

\subsection{Domain Knowledge Injection}
Domain knowledge can also be integrated not only for learning patient state representations but also natural language representations.
~\cite{boag2017awe} enhances the quality of clinical word2vec representation by injecting the knowledge from the UMLS Metathesaurus. 
The authors adopted the concept of dependency-based word embeddings~\cite{levy2014dependency}, which decouples word and context, to learn the domain knowledge-augmented word-level representation using both \texttt{(word, context)} and \texttt{(word, CUI)} pairs.
~\cite{yu2016retrofitting} also learned word representations that are augmented by retrofitting the word vectors built from MeSH terms.

A domain knowledge prior is helpful but not always available since hospitals may use their own in-house terminologies rather than standardized ontology systems, which makes concept mapping difficult.
The bag-of-events method, which utilizes the biomedical concepts and relations in the ontology, is one of the approaches that can standardize the clinical concepts and mitigate such problems~\cite{gong2017predicting}.

\subsection{Cross-domain and Multimodal Representations}
Due to the heterogeneous nature of EHR, representations can also be learned with sources from different domains or modalities in EHR.
~\cite{weng2018mapping,weng2019unsupervised} applied the bilingual dictionary induction algorithm to align two natural language embeddings of different clinical language styles~\cite{conneau2017word}, which are independently trained on non-parallel corpora, and performed cross-domain professional to consumer clinical language translation. 
Such a framework has been proven to be effective even in a cross-modal setting between speech and text corpora~\cite{chung2018unsupervised,chung2018towards}.
Researchers also learned the multimodal representation between image and text for the similar report retrieval task~\cite{hsu2018unsupervised}, as well as for text generation from image input~\cite{liu2019clinically}.
~\cite{weng2019multimodal} further utilized both the multimodal and multitask information to learn a better generalizable representation for pathology metadata prediction problem.

\section{Conclusion}
Representation learning is an important sub-field of machine learning for medicine and healthcare. 
Effective representation learning techniques allow us to obtain models with better performance and interpretability, taking advantage of the complex nature of the data and making do with the limited amount available. 
For a more general survey of using deep learning in EHR, please see~\cite{miotto2017deep} and~\cite{xiao2018opportunities}. 
For medical imaging, we recommend to readers the review about deep learning in medical image analysis~\cite{litjens2017survey}. 

As mentioned in the previous sections, there are several challenges from a machine learning perspective, such as  data insufficiency, temporality and time irregularity, interpretability, and well-utilized cross-domain and cross-modal resources, that we can investigate more to improve the technique of representation learning~\cite{miotto2017deep,xiao2018opportunities,ghassemi2018opportunities}.
There are also other concerns that we should be aware of while learning representations for medicine and healthcare. 
Model biases and fairness are critical issues since the training data we use are usually noisy and biased~\cite{caruana2015intelligible,ghassemi2018opportunities}. 
Model privacy is always a concern that we need to keep in mind and take care of due to emerging techniques of adversarial model attack and model stealing~\cite{tramer2016stealing,madry2017towards}. 
Causality is usually not addressed in most clinical machine learning research, yet it is a critical component for clinical decision making.
~\cite{johansson2016learning} proposed a deep learning framework for counterfactual inference that integrates the ideas of domain adaptation and representation learning.

One step further, we also need to consider how to bring the fruits of research to real products to improve workflow, integrate all information acquired by human and machine, and transform them into clinically actionable solution to improve health outcomes. These are the most important things that we should consider while conducting research on representation learning for medicine and healthcare~\cite{steiner2018impact,chen2018microscope,chen2019develop,sayres2019using}.
This research field still has many unknown properties and deserves more investigation.

\bibliographystyle{icml2019}
\bibliography{weng}

%%%%%%%%%%%%%%%%%%%%%%%%%%%%%%%%%%%%%%%%%%%%%%%%%%%%%%%%%%%%%%%%%%%%%%%%%%%%%%%
%%%%%%%%%%%%%%%%%%%%%%%%%%%%%%%%%%%%%%%%%%%%%%%%%%%%%%%%%%%%%%%%%%%%%%%%%%%%%%%

\end{document}